\def\year{2019}\relax
\documentclass[letterpaper]{article} %DO NOT CHANGE THIS
\usepackage{aaai19}  %Required
\usepackage{times}  %Required
\usepackage{helvet}  %Required
\usepackage{courier}  %Required
\usepackage{url}  %Required
\usepackage{graphicx}  %Required
\usepackage{times}
\usepackage{latexsym}
\usepackage{booktabs}
\usepackage{graphicx}
\usepackage{enumitem}
\usepackage{pgfplots,capt-of}
\usepgfplotslibrary{colorbrewer}
\usepackage{amsmath}
\usepackage{bbm}

\usepackage{url}

\newcommand*\bigcdot{\mathpalette\bigcdot@{2.0}}
\newcommand*\bigcdot@[2]{\mathbin{\vcenter{\hbox{\scalebox{#2}{$\m@th#1\cdot$}}}}}

\begin{document}
\title{REGMAPR - Text Matching Made Easy}
\author{Siddhartha Brahma\\
IBM Research AI, Almaden, USA
}

\maketitle
\begin{abstract}
Text matching is a fundamental problem in natural language processing. Neural models using bidirectional LSTMs for sentence encoding and inter-sentence attention mechanisms perform remarkably well on several benchmark datasets. We propose REGMAPR -- a simple and general architecture for text matching that does not use inter-sentence attention. Starting from a Siamese architecture, we augment the embeddings of the words with two features based on exact and paraphrase match between words in the two sentences. We train the model using three types of regularization on datasets for textual entailment, paraphrase detection and semantic relatedness. REGMAPR performs comparably or better than more complex neural models or models using a large number of handcrafted features.  REGMAPR achieves state-of-the-art results for paraphrase detection on the SICK dataset and for textual entailment on the SNLI dataset among models that do not use inter-sentence attention. 
\end{abstract}

\section{Introduction}
Matching two pieces of text is a common pattern in many natural language processing tasks. For example, in the textual entailment task, given a pair of premise and hypothesis sentences, the task is to classify them into one of three labels \{entailment, contradiction, neutral\} \cite{Bowman2015}. In paraphrase detection, a pair of sentences need to be classified according to whether they are paraphrases of each other \cite{Dolan2005AutomaticallyCA}. In the semantic relatedness task, a pair of sentences need to be scored based on how closely related they are semantically. Other problems like question answering can also be reduced to textual matching by scoring each question-answer pair and picking the answer with the highest score \cite{Wang2017BilateralMM}. In this paper we assume the texts are a pair of sentences $S_1$ and $S_2$. 

There has been a large amount of work on building machine learning models to solve each of these specific problems or text matching in general. In recent years, neural network models have been able to achieve impressive performance on several benchmark  datasets related to these problems. The neural models can be divided into roughly two categories. In the sentence encoder based models, each sentence is encoded into a fixed length distributed representation using a sequence encoder like a BiLSTM \cite{Hochreiter1997} acting on the embeddings of the words in the sentence. The two sentence representations are then composed into a single representation by using heuristic matching features like element-wise difference and element-wise product \cite{Mou2016NaturalLI}, which is then passed through a classification layer. These so-called Siamese architectures are simple, but do not take into account the dependencies between words in the two sentences.
% and also try to  encode all information contained in a sequence into short vectors, which may be very restrictive.
\begin{figure*}[t]
{
\centering
\includegraphics[width=0.9\linewidth]{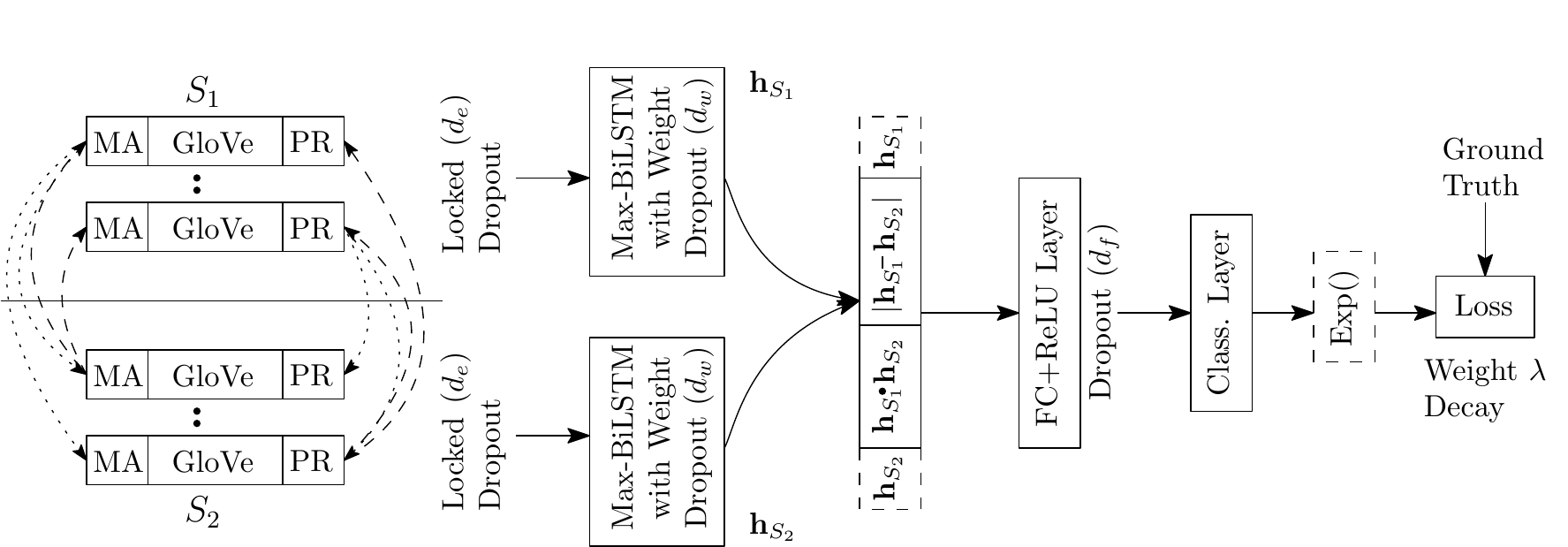}
\caption{Schematic diagram of REGMAPR. The original sentence encodings are used for textual entailment and paraphrase detection and the exponential function is used for semantic relatedness.}
\label{fig:model}
}
\end{figure*}
The other category of neural models incorporate dependencies between words in the two sentences, typically by using an attention mechanism \cite{rocktaschel2016reasoning}. The contextual representation of each word in $S_1$, obtained from the intermediate states of a BiLSTM for example, is composed  with the representations of words in $S_2$ using attention and then compared. This produces a series of representations for words in $S_1$ dependent on the words in $S_2$, which can then be encoded further before being used for classification. Many of the best results in text matching are achieved by architectures that use some form of inter-sentence attention e.g. the ESIM model for textual entailment  \cite{Chen2017}, the BiMPM model for paraphrase detection \cite{Wang2017BilateralMM} and DIIN for both \cite{Gong2018}. While more expressive, these models are quite complex with a large number of parameters. The question remains whether such complex models are absolutely necessary to achieve good performance in text matching problems. In fact, recent work in language modeling has shown that properly regularized vanilla LSTM networks can achieve results that are comparable to the state-of-the-art \cite{Melis2018} without the need for more complex architectures. 

In this paper, we take a middle path and propose a simple Siamese architecture for text matching problems. Each sentence is encoded using a BiLSTM and the representations are composed by computing the element-wise absolute difference and product. Optionally, we also concatenate the original sentence encodings before passing the vector to the classification layer. As mentioned above,  inter-sentence dependence information is crucial for good performance in text matching. To avoid the use of complex attention mechanisms, we augment the embeddings of the words to incorporate inter-sentence information. For each word $t \in S_1$, we add a \emph{matching feature} that indicates whether $t$ appears in $S_2$ to the embedding of $t$. Similarly, for each word  $t \in S_2$, we add a matching feature that indicates whether $t$ appears in $S_1$. Such matching features have been successfully used in neural models for information retrieval \cite{Guo2016ADR}. 
% Note that this kind of matching has been shown to be beneficial in several NLP tasks (CITE).

While the matching feature provides important syntactic information to the model, it is too restrictive. If two words in $S_1$ and $S_2$ that are not exactly same but semantically related, there is a good change that this influences the fact that $S_1$ and $S_2$ are related through an entailment, paraphrase or semantic relationship. In fact, inter-sentence attention mechanisms try to capture some form of semantic dependency by using the contextual representations derived from a BiLSTM. We take a different approach. We use an external database of paraphrase or semantically related words   \cite{Pavlick2015PPDB2B} to capture dependence. For each word $t \in S_1$, we add a \emph{paraphrase feature} to its embedding that indicates whether a paraphrase of $t$ appears in $S_2$. Similarly, for each word  $t \in S_2$, we add a paraphrase feature that indicates whether a paraphrase of $t$ appears in $S_1$.
The matching and the paraphrase features add only two dimensions to the embeddings of each word but capture important syntactic and semantic interaction between the words of the two sentences. 

The importance of regularization in obtaining good generalization performance is a well established fact in deep learning. Several types of regularization specific to recurrent neural networks have been shown to improve performance of LSTM based models e.g. variational dropout \cite{Gal2016ATG} and DropConnect \cite{Merity2018}. We use three types of regularization to train our models in order to achieve good generalization performance.

The base Siamese architecture augmented with the matching and paraphrase features that capture inter-sentence word interaction and regularization define our model -- REGMAPR. We evaluate its performance on six benchmark datasets on textual entailment, paraphrase detection and semantic relatedness. Despite its simplicity, REGMAPR improves upon several existing models which either use complex inter-sentence attention mechanisms or a large number of handcrafted features across all the datasets. It achieves a new state-of-the-art on the SICK dataset for semantic relatedness and on the SNLI dataset for textual entailment among models that do not use inter-sentence attention.

\section{REGMAPR - The Model}
\label{sec:model}
We describe our model by starting from a basic Siamese architecture and augmenting it with additional features. The input to the model is a pair of sentences $S_1$ and $S_2$, with each word mapped to its corresponding distributed representation or word embedding. In this paper we use GloVe embeddings \cite{Pennington2014}). We denote the set of words of $S_i$ by $T(S_i)$ for $i\in\{1,2\}$.

\subsection{BASE}
The basic model uses a standard Siamese architecture. Each sentence is encoded into a single vector using a BiLSTM. As the encoder, we use a max-pooling of the intermediate states of the BiLSTM operating on the sentence. Our choice is inspired by the success of such an encoder in learning general sentence representations \cite{infersent}. In our experiments, we tried other sentence encoders but a max-pooled BiLSTM consistently gave the best results. The encodings of the two sentences $\mathbf{h}_{S_1}$ and $\mathbf{h}_{S_2}$ are composed by concatenating the element-wise absolute difference and element-wise product with the original vectors to form the following feature vector for textual entailment and paraphrase detection.
\begin{equation}
\mathbf{h}_{S_1,S_2} = \left[ \mathbf{h}_{S_1}; \mathbf{h}_{S_2}; |\mathbf{h}_{S_1}-\mathbf{h}_{S_2}|; \mathbf{h}_{S_1}\bigcdot \mathbf{h}_{S_2}\right]
\end{equation}
where \textbf{;} denotes concatenation. Such matching features have been used successfully for textual entailment in the past \cite{Mou2016NaturalLI}. For semantic relatedness, we only use the absolute difference and product, as follows.
\begin{equation}
\mathbf{h}_{S_1,S_2} = \left[  |\mathbf{h}_{S_1}-\mathbf{h}_{S_2}|; \mathbf{h}_{S_1}\bigcdot\mathbf{h}_{S_2}\right]
\end{equation}
%as follows.
%\begin{equation}
%\mathbf{h}_{S_1,S_2} = \left[|\mathbf{h}_{S_1}-\mathbf{h}_{S_2}|; \mathbf{h}_{S_1}\bigcdot\mathbf{h}_{S_2}\right]
%\end{equation}
This feature vector is passed through a fully connected layer, followed by ReLU activation, followed by a classification or scoring layer. 
%For the paraphrase detection task, since it is a binary classification problem, we output a single number which is then binarized using a threshold selected using the Dev set. 
For semantic relatedness, we produce a single number which is then passed through an exponential function and clamped to 1 to constrain it in the range $[0,1]$ \cite{Mueller2016}.  
%The loss function is set to cross-entropy loss for textual entailment and paraphrase detection and mean squared error (MSE) for semantic relatedness. 

\subsection{BASE+REG}
Although regularization is strictly not a part of the architecture, we emphasize its importance. Based on the work of \cite{Merity2018}, we apply three types of regularization.
\begin{enumerate}
\item \emph{Locked Dropout}  ($d_e$)  after the word embedding layer. In this case, a single dropout mask, where each dimension is dropped with probability $d_e$, is selected for a sentence and applied to all the words in the sentence. Also known as variational dropout, its effectiveness has been demonstrated in sequence processing using recurrent neural networks \cite{Gal2016ATG}. 
\item \emph{Dropout} ($d_f$) after the ReLU activation. This is classical dropout proposed in \cite{Srivastava2014DropoutAS}.
\item \emph{Recurrent Dropout}  ($d_w$) on the recurrent weights in the BiLSTM encoder. First proposed in \cite{Wan2013RegularizationON}, this regularization helps reduce overfitting of the hidden-to-hidden weight matrices in the LSTM. 
%\item \emph{L2 weight decay} on the parameters of the model with coefficient  $\lambda$.  
\end{enumerate}

Note that the three types of regularization have been chosen carefully to prevent overfitting in each of the main components of our model. As shown later in the paper, they  help significantly in getting good generalization performance.

%We apply \emph{Locked Dropout} ($d_e$) after the embedding layer and  simple dropout ($d_f$) \cite{Srivastava2014DropoutAS} after the ReLU activation. We also also apply \emph{Weight Dropout}  ($d_w$) on the recurrent weights in the BiLSTM encoder. Finally, we apply a L2 weight decay regularization to the loss term with coefficient $\lambda$. 

\subsection{BASE+REG+MA}
In this model, in addition to the regularization, we augment the word embeddings with a \emph{matching feature}, denoted by MA henceforth. That is, for each word $t$ in $S_i$,  we augment the embedding of $t$ as 
%the embedding $\mathbf{E}_{S_i}(t)$ is augmented as
\begin{equation}
\mathbf{E}_{S_i}^{\text{MA}}(t) = \left[ \text{GloVe}(t); \mathbbm{1}\{t\in T(S_{3-i})\} \right] 
\end{equation}
where $\mathbbm{1}$ is the indicator function. Note that this is a binary feature and provides basic syntactic information to the sentence encoder that the same word is present in both the sentences. 
%Despite its simplicity, it is an excellent signal for text matching problems
%This is repeated for words of $S_2$ as well. We call this the matching or MA feature. 

\subsection{BASE+REG+PR}
In this model, we augment the word embedding with information about the presence of semantically related words in $S_1$ and $S_2$. We use the paraphrase database (PPDB) \cite{Pavlick2015PPDB2B} as the source dictionary of semantically related words. 
%make novel use of the paraphrase database (PPDB) in its lexical form. 
%We obtain a set of pairs of words which are paraphrases of each other.  
For each word $t$ in PPDB, we compute the following set 
\begin{equation}
P(t) = \{ t' | t' \text{ is a paraphrase of } t \text{ in PPDB}\}
\end{equation}
We augment the embedding of $t\in S_i$ using $P(t)$ as follows.
\begin{equation}
\mathbf{E}_{S_i}^{\text{PR}}(t) = \left[ \text{GloVe}(t); \mathbbm{1}\{|P(t)\cap T(S_{3-i})|>0\} \right]  
\end{equation}
%WRITE SOME MORE
We call this the \emph{paraphrase feature} or PR henceforth. This again is a binary feature and is easy to compute once $P(t)$ has been precomputed. 
%This feature provides information to the BiLSTM whether about the semantic relatedness of words in the two sentences. 
Depending on the criteria used for selecting the paraphrase database, this feature can be slightly noisy and yet provides valuable semantic information which is not easily obtainable from the surface forms or word embeddings directly. To the best of our knowledge, this is the first use of PPDB in neural models for text matching.

\subsection{BASE+REG+MA+PR}
The full REGMAPR model combines regularization with the matching and paraphrase features. The word embedding then becomes
\begin{eqnarray}
\mathbf{E}_{S_i}^{\text{MA+PR}}(t) = & \text{[} \text{GloVe}(t); \mathbbm{1}\{t\in T(S_{3-i})\}; \\ \notag
                                                  & \mathbbm{1}\{|P(t)\cap T(S_{3-i})|>0\} \text{]}  
\end{eqnarray}
Note that the full REGMAPR model increases the word embedding by only two dimensions and uses a Siamese architecture for matching the representations of the two sentences. We completely avoid inter-sentence attention mechanisms and instead encode the inter-sentence interaction using the very simple MA and PR features. Crucially, the MA and PR features provide important syntactic and semantic clues that the BiLSTM can exploit. A schematic diagram of the model is shown in Fig. \ref{fig:model}.
REGMAPR is a general architecture which can be applied to any text matching problem. As we show in the next sections, despite its simplicity, it is highly effective in achieving results comparable or better than more complex models. 

\section{Training and Evaluation}

\begin{table}[t]
  \centering
%\resizebox{0.98\columnwidth}{!}{% 
  \begin{tabular}{llll}
    \toprule
     Dataset  & Train & Dev & Test \\
                \midrule
                Textual Entailment \\
                
              	SNLI & 549.4K& 98.4K& 98.2K\\
		SICK-E & 4.5K & 0.5K & 4.9K\\
		\midrule
		Paraphrase Detection \\
		MSRP & 3.7K & 0.4k & 1.7K\\
		QUORA & 384.4K & 10K  & 10K \\
		\midrule
		Semantic Relatedness \\
		
		SICK & 4.5K & 0.5K & 4.9K\\
		STSB & 5.8K &1.5K &1.4K \\
                   \bottomrule
\end{tabular}%
%}
\caption{Datasets and the sizes of the respective train, dev and test sets.}
\label{tab:datasets}
\end{table}

\subsection{Datasets} 
\pgfplotsset{cycle list/Set1-4}
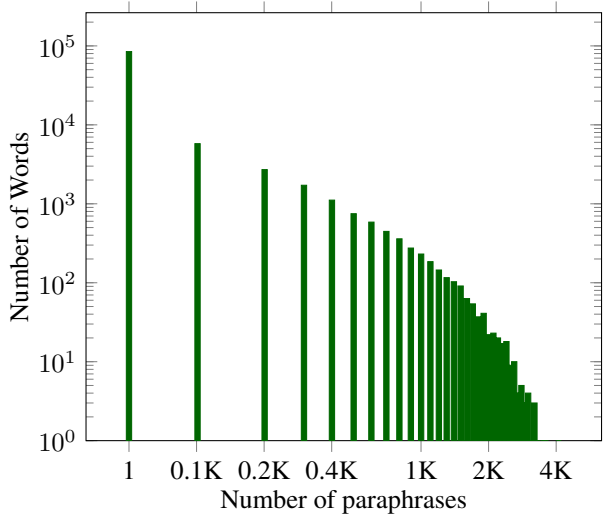
\begin{figure}[t]
%\resizebox{\columnwidth}{!}{%
\begin{tikzpicture}
\usetikzlibrary{patterns}
\begin{loglogaxis}[ybar, bar width=2,
%enlargelimits=0.1,
%x post scale=1.75,
%y post scale=0.7,
xtick={ 50, 100, 200, 400, 1000, 2000, 4000},
xticklabels={ 1, 0.1K, 0.2K, 0.4K , 1K,  2K,  4K},
x tick label style={rotate=0},
xlabel={Number of paraphrases},
ylabel={Number of Words},
y label style={at={(0.05,0.5)}},
ymin=1,
legend cell align={left},
every axis plot/.append style={fill},
]
\addplot[black!60!green]
coordinates
{(50,84640)(101,5776)(201,2709)(301,1717)(401,1113)(501,749)(601,586)(701,447)(801,360)(901,275)(1001,231)(1101,185)(1201,145)(1301,116)(1401,103)(1501,91)(1601,63)(1701,54)(1801,37)(1901,41)(2001,22)(2101,23)(2201,20)(2301,17)(2401,18)(2501,9)(2601,10)(2701,4)(2801,5)(2901,3)(3001,4)(3101,1)(3201,3)(3301,1)(3401,0)(3501,1)(3601,1)(3701,0)(3801,0)(3901,0)(4001,0)(4101,1)(4201,0)(4301,0)(4401,0)(4501,0)(4601,0)(4701,0)(4801,0)(4901,0)};
\end{loglogaxis}
\end{tikzpicture}
%}
\caption{Histogram of words based on the number of paraphrases in PPDB. Both axes are log scaled. The gap between the first two bars has been truncated for ease of visualization.}
\label{fig:ppdb_hist}
\end{figure}
We evaluate our models on six diverse datasets related to three tasks - textual entailment, paraphrase detection and semantic relatedness. For textual entailment, we use the SNLI dataset \cite{Bowman2015} and the SICK-E dataset \cite{Marelli2014ASC}. Each sentence pair in SNLI and SICK-E has a label from the set \{entailment, contradiction, neutral\}. 
 For paraphrase detection we use the MSRP  \cite{Dolan2005AutomaticallyCA} and QUORA  \cite{Quora} datasets . Each sentence pair in these two datasets has a binary label indicating whether they are paraphrases of each other. For semantic relatedness we use the SICK \cite{Marelli2014ASC} and STS Benchmark (STSB)  \cite{Cer2017SemEval2017T1}  datasets, in which each pair of sentences has a semantic relatedness score. The sizes of the train, dev and test sets for each of these datasets is shown in Table \ref{tab:datasets}. 
 
To compute the PR feature, we use the lexical subset of the PPDB paraphrase dataset (specifically the ppdb\_xxl set). There are about 3.7 million pairs of words with an associated score, which we ignore. There are about 99.6K unique words, with more than 50\% having less than 11 paraphrases. A histogram of the frequency of words according to the number of paraphrases is shown in Fig. \ref{fig:ppdb_hist}.
 %We refer readers to the original papers for more details.
 
\subsection{Training} 
For all experiments, we set the LSTM hidden dimension and the dimension of the fully connected layer to 600. We use 300 dimensional GloVe  \cite{Pennington2014} word embeddings. The word embeddings are not updated during training. For the textual entailment and paraphrase detection tasks, a cross-entropy loss function is used, while for the semantic relatedness task the mean squared error (MSE) between the predicted and ground-truth score is used as the loss function. We optimize the weights of the network using  Adam \cite{Kingma2015} with a learning rate of 1e-3, which is decayed by 0.5 when the validation performance drops. 
For each word in PPDB, we construct a one to many map representing its paraphrases. To create the PR feature for a word in a sentence, we lookup this map and check whether there are any words common with the other sentence. 

Hyperparameter search for regularization is done over the following ranges -- locked dropout $d_e \in \{0.0,0.1,0.2,0.3,0.4\}$,  dropout $d_f \in \{0.0,0.1,0.2,0.3,0.4\}$  and  recurrent dropout $d_w \in \{0,0.1,0.2\}$.  
%We vary L2 weight decay $\lambda \in \{0,\text{1e-6, 1e-5}\}$.  
For the SICK dataset, the relatedness scores are linearly scaled from  $[1,5]$ to $[0,1]$. For the STSB dataset, the scores are scaled from $[0,5]$ to $[0,1]$.

%, the relatedness scores are linearly scaled to $[0,1]$ from $[1,5]$ and $[0,5]$, respectively. For these two datasets, we pick the model with the least MSE loss on the dev set. 
%For paraphrase detection, we use the Dev set to select a threshold for binary classification. 

\section{Results}
%We present the performance of our REGMAPR models on the five datasets mentioned above. 
We report results for each of the models defined in Section 2 %\ref{sec:model} 
i.e. BASE, BASE+REG, BASE+REG+MA, BASE+REG+PR and BASE+REG+MA+PR 
for all the six datasets. In all the tables, the best among these is highlighted by bold fonts and the overall best by an underline. 
%Note that, to maintain uniformity, the decimal precision of the numbers vary according to what has been reported in previous publications for the respective problems. 
\subsection{Textual Entailment}
\begin{table}[t]
  \centering
%\resizebox{0.98\columnwidth}{!}{%
  \begin{tabular}{lll}
    \toprule
     Model  & Acc. \\
                     \midrule
                 %DiSAN \cite{Shen2017DiSANDS} & 85.6 \\
                 %Compare Propagate \cite{Tay2018} & 85.9 \\
               % Gumbel Tree LSTM \cite{ChoiGumbel17} & 86.0 \\                
                Shortcut Stacked Encoder \cite{Nie2017ShortcutStackedSE} & 86.0 \\
                %Distance Based Self Attention \cite{Im2017DistancebasedSN} & 86.3 \\
                Reinforced Self Attention \cite{Shen2018ReinforcedSN} & 86.3 \\
                Generalized Pooling \cite{Chen2018EnhancingSE} & 86.6 \\
                \midrule
     %\multicolumn{2}{l}{\textit{REGMAPR (Ours)} }\\
     \midrule
     BASE & 85.2 \\
     BASE+REG & 85.9  \\
     BASE+REG+PR &86.1 \\
     BASE+REG+MA & 86.1\\
     BASE+REG+MA+PR & \underline{\bf{86.8}}  \\
     \bottomrule
\end{tabular}%
%}
\caption{Accuracy results on SNLI test set. Previous results are for models without inter-sentence attention.}
\label{tab:snli_results}
\end{table} 

\begin{table}[t]
  \centering
%\resizebox{\columnwidth}{!}{%
  \begin{tabular}{lll}
    \toprule
     Model  & Acc. \\
                \midrule
                 %DiSAN \cite{Shen2017DiSANDS} & 85.6 \\
              %   Compare Propagate \cite{Tay2018} & 85.9 \\
                Denotational Approach  & 84.6 \\               
                \cite{Lai2014IllinoisLHAD} & \\
                ABCNN \cite{Yin2016ABCNNAC} & 86.2 \\
                Attention Pooling \cite{Yin2017TaskSpecificAP} & \underline{87.2} \\
                    \midrule
     %\multicolumn{2}{l}{\textit{REGMAPR (Ours)} }\\
     \midrule
     BASE & 86.4 \\
     BASE+REG & 86.6  \\
     BASE+REG+PR &86.6 \\
     BASE+REG+MA & 86.7\\
     BASE+REG+MA+PR & \bf{87.0}  \\
     \bottomrule
\end{tabular}%
%}
\caption{Accuracy results on SICK-E test set. }
\label{tab:sicke_results}
\end{table} 
The results of REGMAPR on SNLI are shown in Table \ref{tab:snli_results}.  Regularization helps push the performance of the base model by 0.7\%. Both word matching and paraphrase matching help further, but with a smaller boost of 0.2\%. The combination of MA and PR improves model performance by a much larger 0.9\% compared to the regularized model only. We compare our results with existing models that do not use inter-sentence attention. REGMAPR sets a new state-of-the-art accuracy of 86.8\% for this class of models. Although REGMAPR is not strictly a sentence encoding based model, we do not use sophisticated attention mechanisms like those in  ESIM \cite{Chen2017}.

For the SICK-E dataset, REGMAPR achieves 87.0\% accuracy on the test set, at par with more complex models that use task specific inter-sentence attention mechanisms \cite{Yin2017TaskSpecificAP}. Interestingly, the base model itself achieves 86.4\% accuracy. This points to the fact that simple models with an appropriate number of parameters can sometimes achieve similar or better performance than more complex models, even without regularization. The gain from using regularization and the MA and PR features is modest, maxing out at 0.6\%. One possible reason for this is the relatively small size of the dataset and the class skew in the training set (56\% sentence pairs have neutral labels), as compared to the almost uniform class distribution in the SNLI training set. 

%architecture because computing the MA and PR features requires us to look into both sentences, we do not use any attention mechanism between the sentence encodings e.g. as in  For this class of models, . Investigating the utility of MA and PR features in attention based models will be part of future work. 
\begin{table}[t]
  \centering
%\resizebox{\columnwidth}{!}{%
  \begin{tabular}{lll}
    \toprule
     Model 
                & Acc. & F1 \\
                \midrule
%MT-Metrics & 77.4  & 84.1 \\
 %\cite{Madnani2012ReexaminingMT}  & & \\
TF-KLD \cite{Ji2013DiscriminativeIT}  & 78.6 & 84.6 \\
TF-KLD + Fine-Grained  & \underline{79.9} & \underline{85.4}  \\
 \cite{Ji2013DiscriminativeIT}  & & \\
%TF-KLD (Trans.)   & 80.4 & 85.9 \\
%ConvNet \cite{He2015MultiPerspectiveSS} & 78.6 & 84.7 \\
%TF-KLD-KNN \cite{Yin2015DiscriminativePE} & 78.7 & 84.8 \\
%ABCNN-3 \cite{Yin2016ABCNNAC} & 78.9 & 84.8 \\
Relation Learn & 79.1 & 85.2 \\
\cite{Filice2015StructuralRF} & & \\
     \midrule
   %  \multicolumn{3}{l}{\textit{REGMAPR (Ours)} }\\
     \midrule
%     BASE & 76.8 & 83.3 \\
%     BASE+REG & 77.7 & 83.8 \\
%     BASE+REG+PR & 78.1 & 84.1 \\
%     BASE+REG+MA & 78.6 & 84.5 \\
%     BASE+REG+MA+PR & 79.4 & 85.2  \\
%     
     %BASE & 77.7 & 84.0 \\
      BASE & 77.2 & 84.0 \\
     %BASE+REG & 77.9 & 84.3 \\
          BASE+REG & 77.9 & 84.3 \\
     %BASE+REG+PR & 77.9 & 84.5 \\
     BASE+REG+PR & 78.1 & 84.5 \\
     %BASE+REG+MA & \bf{79.2} & \bf{85.3} \\
     BASE+REG+MA & {78.6} & {84.7} \\
     %BASE+REG+MA+PR & 78.7 & 84.7  \\
     BASE+REG+MA+PR & \bf{79.1} & \bf{85.3}  \\
     \bottomrule
\end{tabular}%
%}
\caption{Accuracy and F1 results on MSRP test set. }
\label{tab:msrp_results}
\end{table} 
\begin{table}[t]
  \centering
%\resizebox{\columnwidth}{!}{%
  \begin{tabular}{lll}
    \toprule
     Model  & Dev & Test \\
                \midrule
                % pt-DECATTword \cite{Tomar2017NeuralPI} & 88.44 & 87.54 \\
                %BIMPM & 88.69 & 88.17 \\
                %\cite{Wang2017BilateralMM}  & & \\
                pt-DECATTchar \cite{Tomar2017NeuralPI} &  88.89 & 88.40\\
                DIIN \cite{Gong2018} & {89.44} & {89.06}\\
                MwAN \cite{Tan2018MultiwayAN} & 89.60 &  \underline{89.12} \\
                \midrule
   %  \multicolumn{3}{l}{\textit{REGMAPR (Ours)} }\\
     \midrule
     %BASE & 87.59 & 87.34 \\
     %BASE+REG & 88.06 & 87.77  \\
     %BASE+REG+PR & 88.30 & 87.94 \\
     %BASE+REG+MA & 88.01 & 87.74\\
     %BASE+REG+MA+PR & 88.87 & 88.62  \\
     BASE & 88.17 & 87.56 \\
     BASE+REG & 88.06 & 88.02  \\
     BASE+REG+PR & 88.37 & 88.12 \\
     BASE+REG+MA & 88.74 & 88.39 \\
     BASE+REG+MA+PR & {89.05} & \bf{88.64}  \\
     \bottomrule
\end{tabular}%
%}
\caption{Accuracy results on the QUORA dev and test set. }
\label{tab:quora_results}
\end{table} 
\begin{table}[t]
  \centering
%\resizebox{\columnwidth}{!}{%
  \begin{tabular}{llll}
    \toprule
     Model 
                & $r$ & $\rho$ &  MSE \\
                \midrule
                   %   Tree-LSTM &  0.8676  & 0.8083 & 0.2532 \\
%\multicolumn{3}{l}{\cite{Tai2015ImprovedSR} }\\
%    ConvNet & 0.8686 & 0.8047 & 0.2606 \\
 %     \multicolumn{3}{l}{\cite{He2015MultiPerspectiveSS} }\\
    Atten. Tree-LSTM & 0.8730 & 0.8117 & 0.2426 \\
    \multicolumn{3}{l}{\cite{Zhou2016ModellingSP} } \\
     Match-LSTM &  0.8822  &0.8345  & 0.2286 \\
    \multicolumn{3}{l}{\cite{Mueller2016} }\\
    IWAN-skip & 0.8833 &  0.8263  & 0.2236 \\
    \multicolumn{3}{l}{\cite{Shen2017InterWeightedAN}}  \\ 
     \midrule
   %  \multicolumn{4}{l}{\textit{REGMAPR (Ours)} }\\
     \midrule
     BASE & 0.8842 & 0.8294 & 0.2224 \\
     BASE+REG & 0.8850 & 0.8311 & 0.2240 \\
     BASE+REG+PR & 0.8857 & 0.8335 & 0.2208 \\
     BASE+REG+MA & 0.8857 & 0.8298& 0.2192 \\
     BASE+REG+MA+PR & \underline{\bf{0.8864}} & \underline{\bf{0.8308}} & \underline{\bf{0.2192}} \\
     \bottomrule
\end{tabular}%
%}
\caption{Pearson correlation ($r$), Spearman correlation ($\rho$) and MSE w.r.t the ground-truth scores on SICK test set.}
\label{tab:sick_results}
\end{table} 
\subsection{Paraphrase Detection}
%We use the same model setup for both the datasets in paraphrase detection. 
The results for the MSRP dataset are shown in Table \ref{tab:msrp_results}.
The trend here is also quite clear with increasing performance as regularization, MA and PR features are added. 
%Surprisingly, the addition of MA helps more than the addition of PR, although this is a paraphrase detection task. This may be an artifact of the small size of the dataset. 
REGMAPR achieves an accuracy of 79.1\%, at par with the results obtained by \cite{Filice2015StructuralRF}, where the authors use, among other things, a combination of more than five handcrafted features (including the MA feature) in a non-neural model. Our model is surpassed only by  \cite{Ji2013DiscriminativeIT}, where the authors use a combination of a term frequency based model and 10 fine grained features. 
%Our model uses simple and general matching features in addition to the neural part. 
%maybe some stats on matching words

The results  on the QUORA dataset are shown in Table \ref{tab:quora_results}. The full REGMAPR model achieves a test accuracy of 88.64\%. This is better than the BiMPM model of \cite{Wang2017BilateralMM} and  pt-DECATTchar  model of \cite{Tomar2017NeuralPI}, both of which use attention mechanisms and, in the latter case, heavy data augmentation. For both the paraphrase datasets, the combination of MA and PR features works the best, reflecting their complementary strengths. 

We emphasize the good performance of REGMAPR on two of the largest datasets (SNLI and QUORA) considered in this paper. More complex models like DIIN \cite{Gong2018} do eventually perform better on both but at the cost of significantly increased model complexity and training time.

\begin{table}[t]
  \centering
%\resizebox{\columnwidth}{!}{%
  \begin{tabular}{lll}
    \toprule
     Model 
                & Dev $r$ & Test $r$ \\
               \midrule
       %          \multicolumn{3}{l}{Feature engineered and mixed} \\
         %        \midrule
%\midrule
  %\multicolumn{3}{l}{Neural Models}  \\
  %\midrule

%SIF \cite{Arora17} & 0.801 & 0.720 \\
  %               Sent2Vec \cite{Pagliardini2018} & 0.787 & 0.755 \\
       %          InferSent \cite{infersent} & 0.801 & 0.758 \\
                 CNN (HCTI) \cite{Shao2017HCTIAS} & 0.834 & 0.784\\
Conversations \cite{YangSTS2018} & 0.835 & 0.808\\ 
%BIT \cite{Wu2017BITAS} & 0.829 & 0.809\\
ECNU \cite{Tian2017ECNUAS} & {0.847} &  \underline{0.810} \\
     \midrule
    % \multicolumn{3}{l}{\textit{REGMAPR (Ours)} }\\
     \midrule
     BASE & 0.797 & 0.786 \\
     BASE+REG & 0.798 & 0.786 \\
     BASE+REG+PR & 0.801 & 0.789 \\
     BASE+REG+MA & 0.802 & 0.793 \\
     BASE+REG+MA+PR & {0.805} & \bf{0.795}  \\
     \bottomrule
\end{tabular}% 
%}
\caption{Dev and test Pearson correlation  ($r$) w.r.t the ground-truth scores on STSB test set.}
\label{tab:sts_results}
\end{table}

\subsection{Semantic Relatedness}
%We use the same model for the two semantic relatedness datasets -- SICK and STS Benchmark. 
For the SICK dataset, REGMAPR sets a new state-of-the-art performance of 0.8864 Pearson correlation, without using any of the additional features or post-processing used by \cite{Mueller2016}. In fact, their base Siamese model built using an LSTM achieves a Pearson correlation  which is about 0.07 less than our base model which achieves 0.8842 Pearson correlation. This points to the significance of using a good sentence encoder.  The improvements obtained from the MA and PR features are significantly lesser than the improvements seen in textual entailment and paraphrase detection. This can be partially explained by the fact that capturing semantic relatedness is a more complex problem.

The results on the STS Benchmark are shown in Table \ref{tab:sts_results}. Here again, the full REGMAPR model performs better than previous neural models like \cite{Shao2017HCTIAS} by almost 1.1\% in Pearson correlation on the test set. Compared to REGMAPR, better performing models like ECNU \cite{Tian2017ECNUAS} use a large number of handcrafted features (66 to be precise) and \cite{YangSTS2018} use transfer learning after training on much larger datasets. 

\pgfplotsset{cycle list/Set1-4}
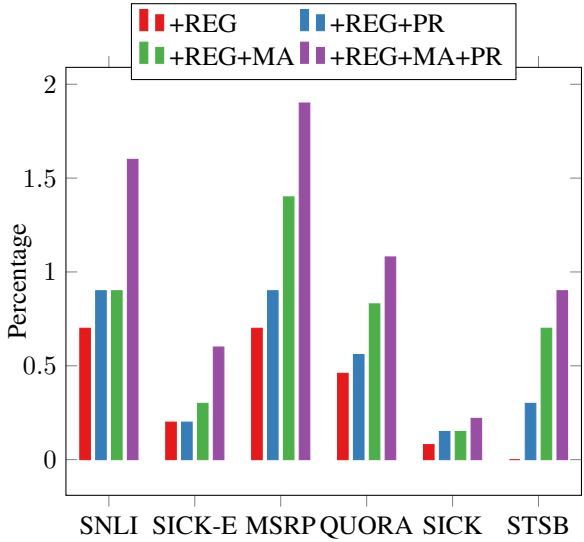
\begin{figure}[t]
%\resizebox{\columnwidth}{!}{%
\begin{tikzpicture}
\usetikzlibrary{patterns}
\begin{axis}[ybar, bar width=4,
enlargelimits=0.1,
%x post scale=1.75,
%y post scale=0.7,
xtick={1, 2,3,4,5,6},
xticklabels={SNLI, SICK-E, MSRP, QUORA, SICK, STSB},
x tick label style={rotate=0},
ylabel={Percentage},
y label style={at={(0.09,0.5)}},
ymin=0,
legend cell align={left},
legend entries={+REG,+REG+PR,+REG+MA,+REG+MA+PR},
legend style={at={(0.5,1.15)}, anchor=north,legend columns=2},
every axis plot/.append style={fill},
cycle list name=Set1-4
]
\addplot
coordinates
	{(1,0.7) (2,0.2) (3,0.7) (4,0.46) (5,0.08) (6,0) };

\addplot
coordinates
	{(1,0.9) (2,0.2) (3,0.9) (4,0.56) (5,0.15) (6,0.3) };

\addplot
coordinates
	{(1,0.9) (2,0.3) (3,1.4) (4,0.83) (5,0.15) (6,0.7) };

\addplot
coordinates
	{(1,1.6) (2,0.6) (3,1.9) (4,1.08) (5,0.22) (6,0.9) };

\end{axis}
\end{tikzpicture}
%}
\caption{Gains in test set performance by using REG, MA and PR  over the BASE model. We use $100\times$Pearson correlation for SICK and STSB and accuracy percentage for others.}
\label{fig:gains}
\end{figure}

\begin{figure}
\begin{tikzpicture}
\begin{axis}[
	width=0.99\linewidth,
       height=0.99\linewidth,
    xlabel={Feature},
    y label style={at={(axis description cs:0.1,.5)},anchor=south},
    ylabel={Relative difference in proportion $R^X$},
    xmin=0.5, xmax=3.1,
    ymin=-0.2, ymax=0.7,
    xtick={1,2,3},
    xticklabels={$\text{X=PR}$ , X=MA, X=MA+PR},
    %tick label style={font=\small},
    %!TEX encoding = UTF-8 Unicodeytick={80, 81, 82, 83, 84, 85, 86, 87},
    legend pos=north west,
    ymajorgrids=true,
    grid style=dashed,
    legend cell align={left},
    legend entries={SNLI, SICK-E, MSRP, QUORA,SICK,STSB}
]
 
\addplot[
    color=blue,
    mark=*,
    ]
    %SNLI
    coordinates {
    (1, 0.049)(2,0.233)(3,0.333)
    };
    \addplot[
    color=magenta,
    mark=*,
    ]
    %SICK-E
    coordinates {
    (1, 0.111)(2,-0.038)(3,-0.167)
    };
    \addplot[
    color=cyan,
    mark=*,
    ]
    %MSRP
    coordinates {
    (1,0.141)(2,0.158)(3,0.333)
    };
    \addplot[
    color=orange,
    mark=*,
    ]
    %QUORA
    coordinates {
    (1,0.0)(2,0.321)(3,0.4)
    };
    \addplot[
    color=black,
    mark=*,
    ]
    %SICK
    coordinates {
    (1,-0.044)(2,0.323)(3,0.444)
    };
    \addplot[
    color=black!30!green,
    mark=*,
    ]
    %SICK
    coordinates {
    (1,0.258)(2,0.305)(3,0.667)
    };
    
 %  \addplot[
   % color=red,
    %mark=square,
    %]
    %coordinates {
    %(512, 82.2)(1024, 83.3)(2048, 84.6)(4096, 85.2)
    %};
    %\legend{SufiSent-Tied, SufiSent, SufiSent-Cat, Sufisent-Cat-Tied,}
 \end{axis}
\end{tikzpicture}
\captionof{figure}{Estimate of predictive power of MA and PR.}
\label{fig:prop}
\end{figure}
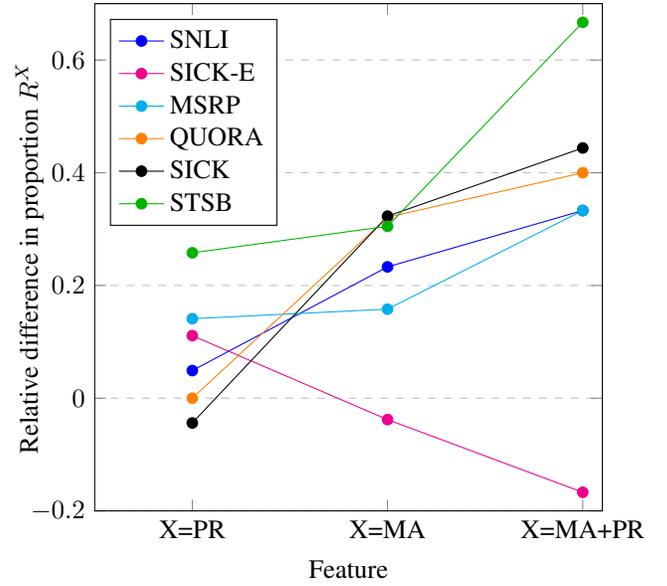
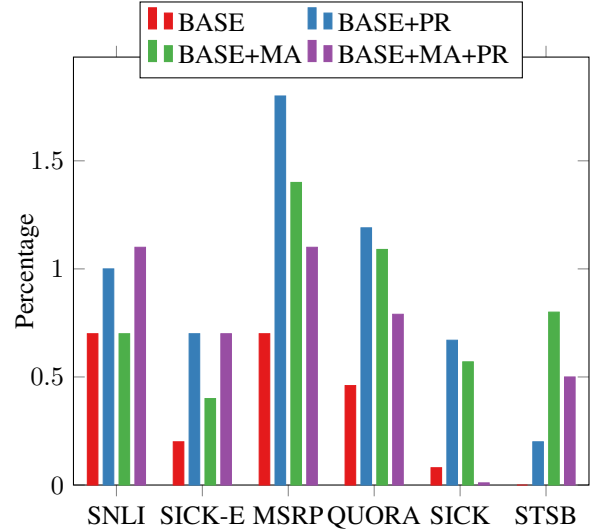
\begin{figure}[h!]
%\resizebox{\columnwidth}{!}{%
\begin{tikzpicture}
\usetikzlibrary{patterns}
\begin{axis}[ybar, bar width=4,
%enlargelimits=0.1,
%x post scale=1.75,
%y post scale=0.7,
xtick={1, 2,3,4,5,6},
xticklabels={SNLI, SICK-E, MSRP, QUORA, SICK, STSB},
x tick label style={rotate=0},
ylabel={Percentage},
y label style={at={(0.09,0.5)}},
ymin=0,
legend cell align={left},
legend entries={BASE,BASE+PR,BASE+MA,BASE+MA+PR},
legend style={at={(0.5,1.13)}, anchor=north,legend columns=2},
every axis plot/.append style={fill},
cycle list name=Set1-4
]

\addplot
coordinates
	{(1,0.7) (2,0.2) (3,0.7) (4,0.46) (5,0.08) (6,0) };

\addplot
coordinates
	{(1,1.0) (2,0.7) (3,1.8) (4,1.19) (5,0.67) (6,0.2) };

\addplot
coordinates
	{(1,0.7) (2,0.4) (3,1.4) (4,1.09) (5,0.57) (6,0.8) };

\addplot
coordinates
	{(1,1.1) (2,0.7) (3,1.1) (4,0.79) (5,0.01) (6,0.5) };

\end{axis}
\end{tikzpicture}
%}
\caption{Gains in test set performance by using REGularization over BASE, BASE+MA, BASE+PR and BASE+MA+PR.}
\label{fig:reg_gains}
\end{figure}

\section{Analysis of Results}
In this section, we analyze the results presented in the previous section by comparing the contribution of each of the main components of REGMAPR. In Fig.~\ref{fig:gains}, we summarize the gains in performance due to each component of REGMAPR over the base model for all the six datasets. Some trends can be ascertained from the plot. In all the cases, the MA feature helps equally or more than the PR feature. The combination of MA and PR consistently produces the highest gains, which justifies our choice of modeling inter-sentence word interaction using both of these features.

To further illustrate the effect of the two features, we investigate the correlation of the the presence of these features with the class labels or scores. For each of the datasets, we partition the training set into two classes - positive ($P$) and negative ($N$). For MSRP and QUORA, pairs of sentences that are paraphrases are positive  and pairs that are not paraphrases are  negative. For SNLI and SICK-E, pairs of sentences that have the entailment label are positive and those that have the contradiction label are negative. For SICK and STSB, we compute the average semantic relatedness score in the training set and mark all pairs that have score greater than equal to the average as positive and the remaining as negative. Next, over all the pairs in $P$, we compute the proportion of words that have the MA feature on as follows.
\begin{equation}
R_P^{\text{MA}} = \frac{{\sum_{(S_1,S_2)\in P}} |\{t | \mathbf{E}_{S_1}^{\text{MA}}(t)=1\}| +  |\{t | \mathbf{E}_{S_2}^{\text{MA}}(t)=1\}|  }{{\sum_{(S_1,S_2)\in P}} |S_1|+|S_2|}
\end{equation}
Similarly, we compute the proportions $R_N^{\text{MA}}, R_P^{\text{PR}}, R_N^{\text{PR}}$ and $R_P^{\text{MA+PR}}, R_N^{\text{MA+PR}}$. For $\text{MA+PR}$, we count the words that have both the features on. These proportions may be interpreted as estimates of how likely a word occurring in a sentence pair with a positive (or negative) label has the MA, PR or both features on. Finally, to estimate the predictive power of the features, we condense the ratios for the positive and negative classes into one number as follows. 
\begin{equation}
R^{\text{MA}} = \frac{R_P^{\text{MA}}- R_N^{\text{MA}}}{(R_P^{\text{MA}}+R_N^{\text{MA}})/2}
\end{equation}
The ratios $R^{\text{MA}}$ and $R^{\text{MA+PR}}$ are defined similarly.  A higher positive value of this ratio means that it is more likely that the corresponding feature is present in words occurring in sentences in $P$ as compared to $N$ and hence can have higher predictive power. 

We plot the values of $R^{\text{MA}}$, $R^{\text{PR}}$ and $R^{\text{MA+PR}}$ for each of the six datasets in Fig. \ref{fig:prop}. Except for the SICK-E dataset, there is a consistent increase as we move from $R^{\text{PR}}$, to $R^{\text{MA}}$ and finally to $R^{\text{MA+PR}}$. This partially explains the relative gains shown in Fig. \ref{fig:gains}. Since we used the PPDB database without any filtering, there may be noisy paraphrases resulting in values of $R^{\text{PR}}$ that are close to zero for some datasets.

Finally, to evaluate the effect of regularization (REG), we plot the gains obtained by including it in the presence or absence of the MA and PR features in Fig. \ref{fig:reg_gains}. It is clear that proper regularization helps a great deal in improving generalization and hence test set performance. The trends across the features here are less clear, although regularization tends to help most when only the PR feature is used. 

\section{Related Work}
Text matching is a general problem in NLP and the three specific tasks considered in this paper have a long history of their own. For textual entailment, 
%the availability of large datasets like SNLI \cite{Bowman2015} has spurred a lot of research in the past few years. 
most of the state-of-the-art architectures for the SNLI dataset use inter-sentence attention mechanisms, canonical examples being ESIM \cite{Chen2017} and BiMPM \cite{Wang2017BilateralMM}. The attention  in these models is typically computed using the intermediate states of a BiLSTM. On the other hand, we encode the word interaction features in the word embeddings and are not parameter free. Purely Siamese architectures have been used in the past for textual entailment \cite{Mou2016NaturalLI},  but they have lower accuracy. 

For paraphrase detection, many of the best models for the MSRP dataset use a combination of handcrafted features in a non-neural setting. These include KL-Divergence based features in \cite{Ji2013DiscriminativeIT} and syntactic and semantic matching features in \cite{Filice2015StructuralRF}. The latter work uses at least five features, including the MA feature. REGMAPR uses fewer and simpler features and incorporates them in a neural model to achieve comparable performance. On the larger QUORA dataset, most existing models use inter-sentence attention mechanisms. %, sometimes combined with large scale pretraining \cite{Tomar2017NeuralPI}. 

Previous work on semantic relatedness has centered largely on combining neural models with handcrafted features. The state-of-the-art model for STSB \cite{Tian2017ECNUAS} combines a total of 66 features with a neural model. Our model achieves comparable performance with far lesser complexity. The use of the exponential function for semantic relatedness scoring was first introduced by \cite{Mueller2016} who apply it directly to the L1 distance between the encodings of the two sentences. Our model uses a more complex matching function followed by fully connected layers before applying the exponential function. 

In a related work, the authors in  \cite{Tymoshenko2015AssessingTI} devise a number of syntactic and semantic matching features for the answer passage reranking task from information retrieval in a non-neural setting. The syntactic features include the MA feature and the semantic matching features are derived from external resources like YAGO, DBPedia and Wikipedia. In our model, we use PPDB as the only external resource of semantic matching information and let the neural network learn the features that are most suitable for a particular task. 
%We are not aware of any other work that uses the PPDB for classification of sentences or sentence pairs. 

Finally, there has been some work on exploring general architectures for text matching problems. The work of \cite{Wang2016ACM} explores various techniques for estimating word interaction through an attention mechanism. Similar approaches were explored in \cite{He2016PairwiseWI}, \cite{Parikh2016ADA} and  \cite{Wang2016MachineCU}. Among more recent models, \cite{Gong2018} and \cite{Kim2018SemanticSM} have achieved state-of-the-art performance on the SNLI and QUORA datasets using multi-layered and highly complex attention mechanisms. 

The use of dropout regularization for better generalization is a well established principle in deep learning. Our work adopts a subset of the suite of regularizations successfully used in language modeling by \cite{Merity2018}. Of particular significance is the use of recurrent dropout or DropConnect \cite{Wan2013RegularizationON} for the weights of the BiLSTM. To the best of our knowledge, ours is the first work to explore its use in text matching problems. 

\section{Conclusion}
In this paper, we propose REGMAPR -- a neural model for text matching that incorporates simple word interaction features in a Siamese architecture and train it with three different types of regularization. Our model performs comparably or better than many existing models that use complex inter-sentence attention mechanisms or many handcrafted features on six diverse datasets. In future work, we plan to explore further ways to infuse inter-sentence semantic information in the word embeddings and matching heuristics over multi-layered sentence representations. 
%Earlier work that use We do not use such attention mechanisms 
%Each task considered in this paper are important problems in NLP. While neural models often give state-of-the-art results, many of them have complex architectures. On the other hand, the work of \cite{Melis2018}  and \cite{Merity2018} has shown that properly regularized standard LSTM models give comparable or better results than more complex models. Our work is most closely related to \cite{Wang2016ACM}, where the authors use a  carefully constructed word interaction model which is then passed through a convolutional layer. We obtain comparable or better results using a simple Siamese architecture augmented by matching and paraphrase features for each word and suitable regularization.
%
%\cite{Tymoshenko2015AssessingTI} - syntactic and semantic feature matching but non neural using complex wikipedia categories, etc. 
%\cite{Filice2015StructuralRF} - lots of feature matching for question pairs

%On theOur model is a simple Siamese architecture \cite{Mou2016NaturalLI} and simple word and paraphrase matching features. 

%\section{Conclusion}
\bibliographystyle{aaai}
\bibliography{bibtex}

\end{document}